\documentclass[10pt,twocolumn,letterpaper]{article}

\usepackage{iccv}
\usepackage{times}
\usepackage{epsfig}
\usepackage{graphicx}
\usepackage{amsmath}
\usepackage{amssymb}
\usepackage{multirow}
\usepackage{mathtools}

\usepackage[breaklinks=true,bookmarks=false]{hyperref}

\iccvfinalcopy 

\def\iccvPaperID{22} 

\ificcvfinal\fi

\begin{document}

\title{SNIDER: Single Noisy Image Denoising and Rectification for \\Improving License Plate Recognition}

\author{Younkwan Lee \qquad Juhyun Lee \qquad Hoyeon Ahn \qquad Moongu Jeon\\
Machine Learning and Vision Laboratory\\
Gwangju Institute of Science and Technology (GIST), Korea\\
{\tt\small\{brightyoun, leejuhyun, ajhoyeon, mgjeon\}@gist.ac.kr}
\and
}

\maketitle
\ificcvfinal\thispagestyle{empty}\fi

\begin{abstract}
   In this paper, we present an algorithm for real-world license plate recognition (LPR) from a low-quality image. Our method is built upon a framework that includes denoising and rectification, and each task is conducted by Convolutional Neural Networks. Existing denoising and rectification have been treated separately as a single network in previous research. In contrast to the previous work, we here propose an end-to-end trainable network for image recovery, Single Noisy Image DEnoising and Rectification (SNIDER), which focuses on solving both the problems jointly. It overcomes those obstacles by designing a novel network to address the denoising and rectification jointly. Moreover, we propose a way to leverage optimization with the auxiliary tasks for multi-task fitting and novel training losses. Extensive experiments on two challenging LPR datasets demonstrate the effectiveness of our proposed method in recovering the high-quality license plate image from the low-quality one and show that the the proposed method outperforms other state-of-the-art methods.
\end{abstract}

\section{Introduction}
   License plate recognition (LPR) from the real-world is one of the fundamental problems in several intelligent transport systems (ITS) applications such as vehicle re-identification \cite{liu2016deep,shen2017learning}, outdoor scene understanding \cite{cherng2009critical,noh2012new}, and de-identification for privacy protection \cite{du2011preservative}. In the last few years, LPR has been widely studied in theoretical, experimental and numerical ways to provide robust image representation. Many LPR methods \cite{bulan2017segmentation,anagnostopoulos2006license,gou2015vehicle,li2016reading} are capable of capturing the structural properties of images and noise for carefully constrained settings. Despite the recent success, recognizing license plate in the wild is still far from satisfactory due to the variations that suffer from appearance, noise, angle, and illumination.

    \begin{figure}[t]
    \begin{center}
       \includegraphics[width=1.0\linewidth]{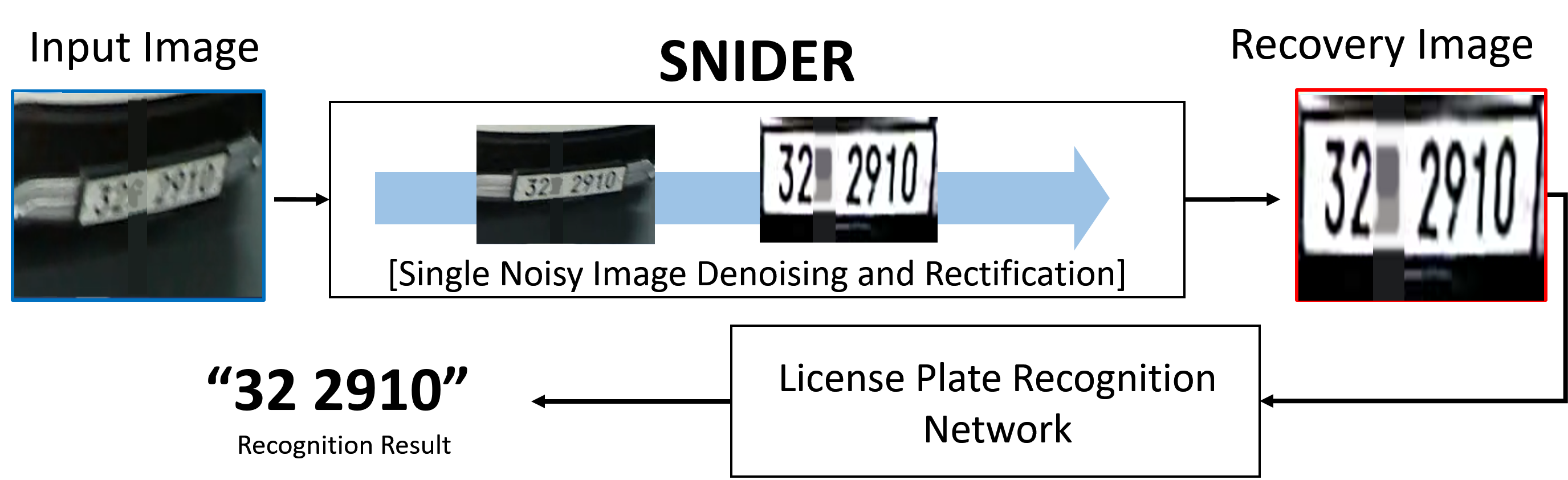}
    \end{center}
       \caption{The proposed system consists of two components: single noisy image denoising and rectification (SNIDER) for recovering a low-quality license plate image and a license plate recognition (LPR) network for recognizing the final recovery image. The SNIDER is an end-to-end trainable network with auxiliary tasks for better image recovery. The LPR network uses a pre-trained DarkNet based on YOLO v3 to detect texts.}
    \label{fig:long}
    \label{fig:onecol}
    \end{figure}
   
   Recently, due to the hierarchical feature extraction and learning capability, deep convolutional neural networks (CNNs) have made remarkable advances in many computer vision applications, such as object detection \cite{ren2015faster,redmon2018yolov3}, semantic segmentation \cite{long2015fully,ronneberger2015u}, action recognition \cite{yu2018driver}, and face recognition \cite{wen2016discriminative,park2017learning}. As a result, CNN-guided LPR methods are also extensively applied to handle the problem of recognizing license plate captured directly real-world camera. For example, Zhuang \textit{et al.} \cite{zhuang2018towards} transform license plate into a semantic segmentation result with the counting network to handle appearance variations. Although numerous LPR methods have been developed \cite{silva2018license,zhuang2018towards}, they are not still capable of learning all types of samples in the wild. For this reasons, their algorithms practically assume a high-quality image as an input. Generally, the typical appearance of the license plate collected in real-world scenes might contain the aforementioned challenges, causing deterioration in LPR performance. Hence, developing and implementing robust LPR framework are highly indispensable, especially for real-world scenes.

   In this paper, we design an end-to-end single noisy image denoising and rectification network (SNIDER) for better LPR based on multiple auxiliary tasks. Figure 1 illustrates the LPR framework in which the proposed SNIDER is combined with a pre-trained LPR network. The SNIDER consists of two sub-networks: a denoising network and a rectification network. Motivated by the success of U-Net \cite{ronneberger2015u} in recovering the object details, we employ U-Net structure as an image recovery backbone network, attempting to extract visual content at structural-level details. In the denoising sub-network (DSN), we try to transform a low-quality image to a high-quality image pixel by pixel directly. The DSN can penalize the loss between noisy and noise-free image pairs and thus acquire the output image with the fine textures of the clean component, learning an independent realization of the noise. However, even with such sophisticated DSN, denoising images are unsatisfactory because they still have arbitrary geometric variations. Therefore, the rectification sub-network (RSN) is proposed to correct geometric distortions of denoising license plates and generate more accurate correction image distortion. Furthermore, we propose to leverage the new auxiliary tasks to further optimize the image recovery sub-networks (DSN, RSN) of SNIDER. There are two auxiliary tasks: a text counting module and a segment prediction module. Specifically, we solve each auxiliary module using CNN as a decoder. The counting module is used to predict the number of text in the image as a classification problem. In this module, despite the ambiguous boundary of consecutive text, text counting can distinguish single text, which makes the image quality suitable for text detection. For the segment prediction module, we propose a binary segmentation to emphasize the foreground over the background. The generated segmentation result makes the license plate clean for text recognition. Finally, learning the auxiliary tasks will lead the intermediate features of the recovery main task networks to enhance the difficulties such as geometric variations and low-quality information. More importantly, we introduce a new loss function that trains the SNIDER with auxiliary tasks, which provide significantly higher license plate quality for robust LPR.
   
   To sum up, we highlight the main contributions of this paper as follows: 
   \begin{itemize}
       \item We propose a novel end-to-end license plate recovery network, where denoising and rectification network are used to generate a clear recovery image for robust LPR performance.
       \item We present the auxiliary tasks to leverage the quality of the license plate recovery from low-quality. Mainly a new loss is introduced to provide regularization effects to the backbone SNIDER for robust representation and license plate recovery.
       \item Finally, we demonstrate the effectiveness of the proposed method in recovering a high-quality license plate from a low-quality license plate in the real-world and show that the LPR performance outperforms the state-of-the-art methods on two challenging datasets, AOLP-RP \cite{hsu2012application} and VTLPs dataset newly collected on the most challenging real-world environments.
   \end{itemize}

    \begin{figure*}[t]
    \begin{center}
       \includegraphics[width=1.0\linewidth]{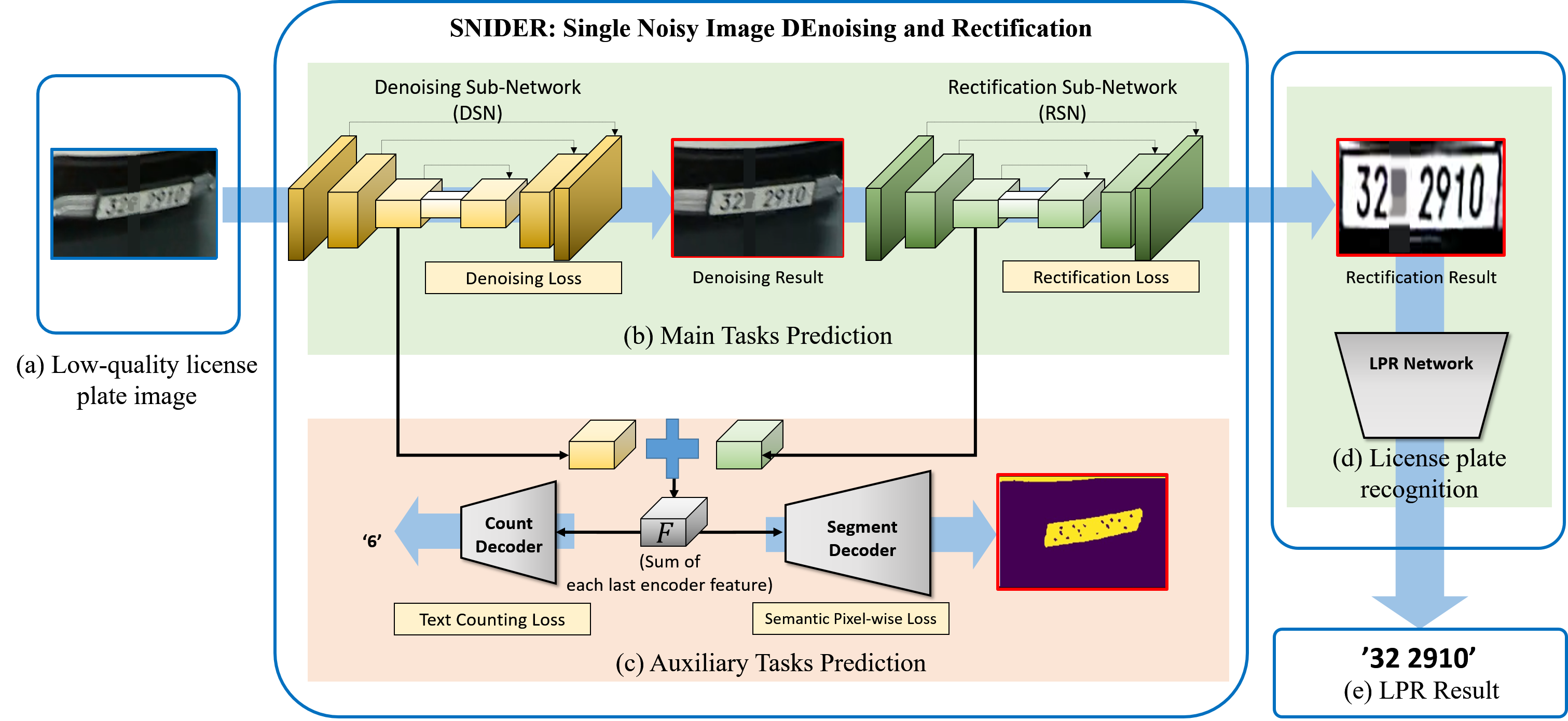}
    \end{center}
       \caption{The training and testing process of the proposed approach with the learning of two auxiliary tasks: (a) The input images are fed into SNIDER for the image recovery; (b, c) SNIDER consists of main tasks (\textit{i.e.} DSN, RSN) and auxiliary tasks, they transform low-quality data into high-quality data for training the DSN, RSN and auxiliary tasks networks; (d, e) LPR network is for testing and outputs a LPR result. The DSN is trained to generate a denoising image from the low-quality input image. Also, the RSN is trained to generate a rectified image from the result of DSN. The auxiliary tasks include text counting and binary segmentation, which are formulated as classification and regression simultaneously.}
    \label{fig:long}
    \label{fig:onecol}
    \end{figure*}

\section{Related Work}
In this section, we briefly review on low-quality image recovery methods and license plate recognition methods that is most related to this work.

\subsection{Low-Quality Image Recovery}
To obtain the high-quality image, most of the existing methods depend on the assumption that both signal and noise arise from particular statistical regularities by using hand-crafted algorithms, such as anisotropic diffusion \cite{perona1990scale} and total variation \cite{rudin1992nonlinear}. Besides, non-parametric models \cite{dabov2007video,pan2016blind} were developed to model image noise, but they were also not robust to the unconstrained environment in the wild due to priors estimated from limited observations. Recently, due to the advances in deep learning, most denoising algorithms are designed with deep neural network architectures and data-driven approach rather than relying on the priors. Burger \textit{et al.} \cite{burger2012image} employ multi-layer perceptrons with a data-driven technique based on an extensive image database. Zhang \textit{et al.} \cite{zhang2017beyond} train the deep CNN by utilizing batch normalization (BN) \cite{ioffe2015batch} and residual learning \cite{he2016deep}.

Though useful for estimating a clean image, text classifiers are still hard to recognize due to the irregular text geometry. It motivates research for image recovery to extend image rectification. Shi \textit{et al.} \cite{shi2016robust} develop a spatial transformer network (STN) for rectifying text distortion. Cheng \textit{et al.} \cite{Cheng_2017_ICCV} adopt more in-depth representations of images by a residual network. Different from the existing methods, in this paper, we extract deep representations of images using the U-Net-based CNN for denoising as well as rectification. To the best of our knowledge, our research may be first work to apply the two modules mentioned above for LPR at the same time. 

\subsection{License Plate Recognition}
Before the advent of deep learning, most of the traditional LPR methods \cite{kim2000learning,anagnostopoulos2006license,hsu2012application,zhu2015end} employ two-stage process flow, involving text detection and following text recognition. After the advancement of deep learning, many approaches employ a one-stage process flow without text detection.  Li \textit{et al.} \cite{li2016reading} extract deep feature representations by using RNN with LSTM for acquiring sequential features of the license plate. Bulan \textit{et al.} \cite{bulan2017segmentation} estimate domain shifts between target and multiple source domains for selecting a domain that yields the best recognition performance based on fully convolutional network \cite{long2015fully}. However, these methods only consider high-quality license plate image except for low-quality image, which is easily led to low performance in real-world scenes. Moreover, their methods lack little or no effort to improve image quality, while requiring a lot of computing power. In this work, unlike existing methods, we adopt image recovery for high LPR performance under the low-quality image in real-world scenes. To the best of our knowledge, this is the first time we apply sophisticated image recovery to handle a challenging real-world environment. Besides, our methods are computationally efficient and capable of real-time recognition despite additional recovery modules.

    \begin{figure}[t]
    \begin{center}
       \includegraphics[width=1.0\linewidth]{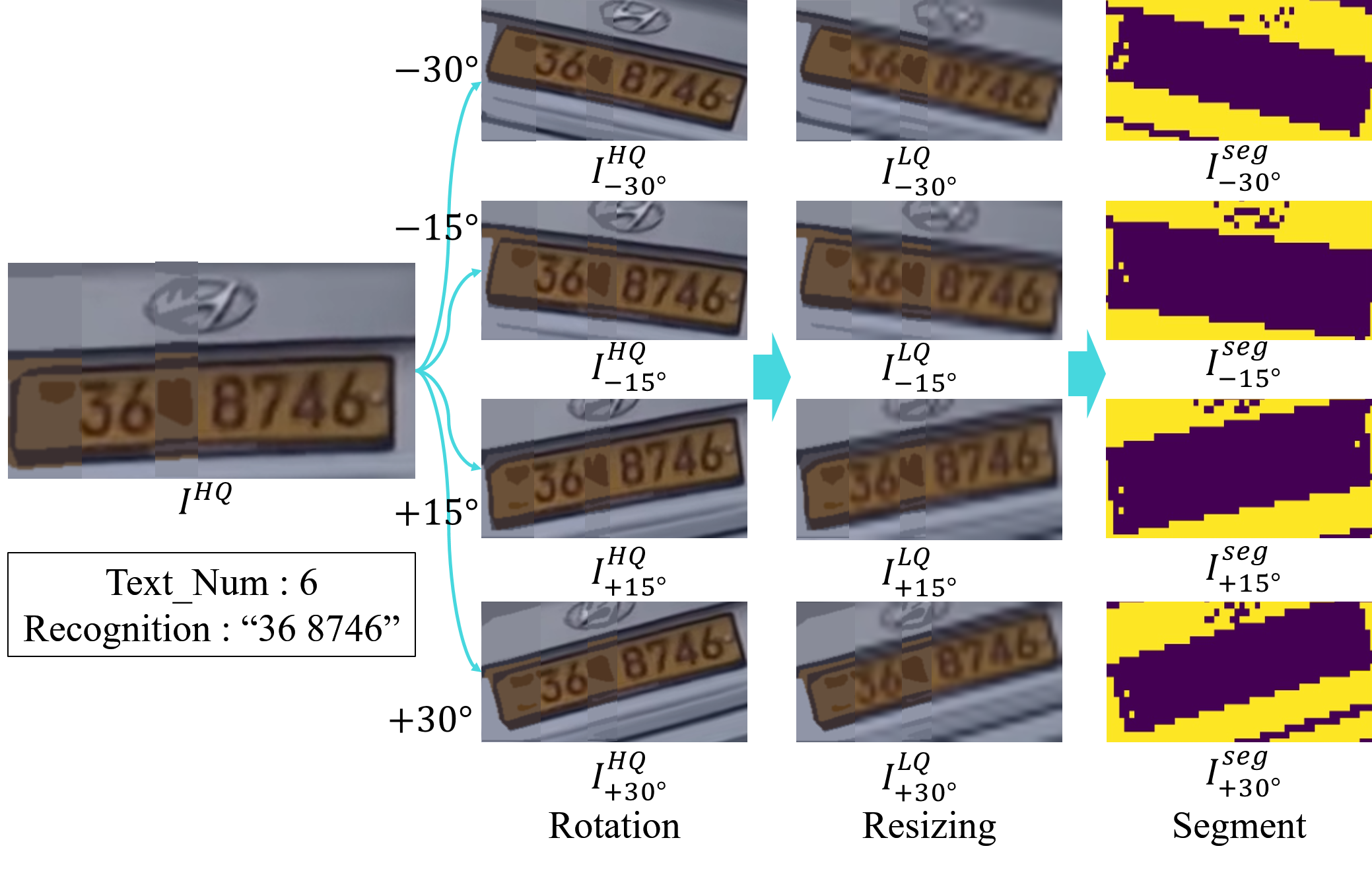}
    \end{center}
       \caption{Label generation for the training of the proposed method. From a high-quality image as ground truth, the rotated images can be obtained using a simple linear transformation, and the low-quality image is processed through downsampling $\times$ 1/4 of rotation image. The segmented image is inferred through binarization \cite{cai2014new} of the low-quality image.}
    \label{fig:long}
    \label{fig:onecol}
    \end{figure}

\section{Proposed Method}
The proposed approach consists of three parts: 1) main tasks prediction networks $G_D$ and $G_R$ for denoising and rectification; 2) auxiliary tasks prediction networks ${D_c}$ and ${D_s}$ for count classification and segment prediction; 3) LPR network for text detection and classification. The proposed architecture is illustrated in Figure 2. For training, dataset for main tasks and auxiliary tasks can be inferred from the intentionally transform operation by simple rotation (for rectification) and down-resizing (for denoising), as shown in Figure 3. In particular, only one sample of original image $I^{HQ}$ simply can generate four training samples that have been transformed by different angles. Given the training samples $I^{HQ}_{i}$ for $G_D$, $I^{LQ}_{i}$ for $G_R$, $I^{seg}_{i}$ for ${D_s}$ and $c$ for ${D_c}$,  $i \in \{{-30\,^{\circ}},{-15\,^{\circ}},{+15\,^{\circ}},{+30\,^{\circ}} \}$, the main tasks $G_D$ and $G_R$ extract recovery result from input image $I^{LQ}_{i}$ and corresponding samples. LPR network $LPR$ then takes $G_R(G_D(I^{LQ}_{i}))$ to recognize a recovery image.

In the following subsections, we introduce the method to predict the main tasks in Section 3.1. Then, we also address the auxiliary tasks for prediction in Section 3.2. Then, we describe the network training of the proposed architecture in Section 3.3. Finally, we illustrate the testing process in Section 3.4.

\subsection{Denoising and Rectification Network}
Our main task networks include two sub-networks (\textit{i.e.} denoising sub-network and rectification sub-network), and the first sub-network takes the low-quality image as the input, and the output is the recovered image. In this paper, we design the rectification network to rectify the denoising results from the denoising network. 

The image recovery results \cite{Isola_2017_CVPR} have shown the effectiveness of the U-Net since it can provide high-quality overall details of an image object, without a negative impact on the image generation. Therefore, we adopt a U-Net-based architecture adding skip connections that shuffle low-level information shared between input and output across the network. In contrast to their network, our recovery network includes two sub-networks, which are also the U-Net architecture. As shown in Table 1 and Figure 2.(b,c), our denoising network $G_D$ and rectification network $G_R$ consist of the encoder and decoder module.

To achieve the main tasks, we first feed $I^{LQ}_{i}$ into $G_D$ to generate denoising results. Given a pair of input image and non-rectified ground-truth denoising image $\{{I^{LQ}_{i,j}, I^{HQ}_{i,j}}\}^N_{(i,j)}$, loss function for the $G_D$ is the pixel-wise MSE loss, and it is calculated as Eq. (1):

    	\begin{equation}
    	\begin{split}
        	\mathcal{L}_{G_D}(w) = { \frac{1}{N}\sum_{(i,j) \in N}^{N}\parallel G_{D_w}(I^{LQ}_{(i,j)}) - I^{HQ}_{(i,j)} \parallel^2},
    	\end{split}
    	\end{equation} where $w$ is the parameters of denoising network. Such loss function encourages the $G_D$ to not only extract the content information of input image but also generate a high-quality natural image in pixel level.

Then, the rectification sub-network $G_R$ processes the output from $G_D$, and outputs a rectified high-quality image, which is easier for the LPR network to recognize the identification text. With the training pairs of $\{{G_D(I^{LQ}_{i,j}), I^{HQ}_{i=0,j}}\}^N_{(i,j)}$, the $G_R$ can be trained using a L1 loss for the predicted result $G_R(G_D(I^{LQ}_{i,j}))$:

    	\begin{equation}
    	\begin{split}
        	\mathcal{L}_{G_R}(w) = { \frac{1}{N}\sum_{(i,j) \in N}^{N}\parallel G_{R_w}(G_{D_w}(I^{LQ}_{(i,j)})) - I^{HQ}_{(i=0,j)} \parallel^1},
    	\end{split}
    	\end{equation} where $w$ is the parameters of the rectification network. Unlike L2 loss, using L1 loss in the pixel level helps to preserve the appearance of an object, such as image color, intensity, and illumination, and leads to denoising result capable of only geometric transformation. Therefore, we can only perform geometric transformations without the appearance damage of the image during the rectification process, which forces the recognizer to be helpful.

\begin{table}[]
\begin{center}
\caption{Details of different proposed network architectures.}
\begin{tabular}{l|l|l}
\hline
\scriptsize{}                                                                                              & \multicolumn{1}{c|}{\scriptsize{SNIDER}}                                                                                                                                                                                                                                                                                                                                                                                                                         & \scriptsize{SNIDER-Tiny}                                                                                                        \\ \hline
\begin{tabular}[l]{@{}l@{}}
\scriptsize{Encoder} \\ 
\scriptsize{(for $G_D$,} \\
\scriptsize{$G_R$)}\end{tabular} 
& \begin{tabular}[c]{@{}l@{}}
\scriptsize{conv1(1,2) : (3$\times$3$\times$32 conv)$\times$2, stride 2}\\ 
\scriptsize{pool1 : (2$\times$2) max pooling, stride 2}\\ 
\scriptsize{conv2(1,2) : (3$\times$3$\times$64 conv)$\times$2, stride 2}\\ 
\scriptsize{pool2 : (2$\times$2) max pooling, stride 2}\\ 
\scriptsize{conv3(1,2) : (3$\times$3$\times$128 conv)$\times$2, stride 2}\\ 
\scriptsize{pool3 : (2$\times$2) max pooling, stride 2}\\ 
\scriptsize{conv4(1,2) : (3$\times$3$\times$256 conv)$\times$2, stride 2}\\ 
\scriptsize{pool4 : (2$\times$2) max pooling, stride 2}\\ 
\scriptsize{conv5(1,2) : (3$\times$3$\times$512 conv)$\times$2, stride 2}\end{tabular} & 
\begin{tabular}[c]{@{}l@{}}
\scriptsize{7$\times$7$\times$32 conv,}\\ \scriptsize{stride 2}\\ 
\scriptsize{7$\times$7$\times$64 conv,}\\ \scriptsize{stride 2}\\ 
\scriptsize{5$\times$5$\times$128 conv}\end{tabular}   \\ \hline

\begin{tabular}[l]{@{}l@{}}
\scriptsize{Decoder} \\ 
\scriptsize{(for $G_D$,} \\ \scriptsize{$G_R$,} \\ \scriptsize{$D_s$)}\end{tabular} & \begin{tabular}[c]{@{}l@{}}
\scriptsize{x2 upsample : concat(conv5\_2, conv4\_2)}\\ 
\scriptsize{conv6(1,2) : (3$\times$3$\times$256), stride 2}\\ 
\scriptsize{x2 upsample : concat(conv6\_2, conv3\_2)}\\
\scriptsize{conv7(1,2) : (3$\times$3$\times$128), stride 2}\\ 
\scriptsize{x2 upsample : concat(conv7\_2, conv2\_2)}\\ 
\scriptsize{conv8(1,2) : (3$\times$3$\times$64), stride 2}\\ 
\scriptsize{x2 upsample : concat(conv8\_2, conv2\_2)}\\ 
\scriptsize{conv9(1,2) : (3$\times$3$\times$32), stride 2}\\ 
\scriptsize{conv10 : (1$\times$1$\times$3)}\end{tabular}                      & 
\begin{tabular}[c]{@{}l@{}}
\scriptsize{5$\times$5$\times$64 conv}\\ 
\scriptsize{7$\times$7$\times$32 deconv,} \\\scriptsize{dilate 2}\\ 
\scriptsize{7$\times$7$\times$3 deconv,} \\\scriptsize{dilate 2}\end{tabular} \\ \hline

\begin{tabular}[l]{@{}l@{}}
\scriptsize{Decoder} \\ 
\scriptsize{(for $D_c$)}\end{tabular}                                    & \multicolumn{1}{c|}{\begin{tabular}[c]{@{}l@{}}
\scriptsize{N$\times$N$\times$512 conv}\\ 
\scriptsize{1$\times$1$\times$256 conv}\\ 
\scriptsize{1$\times$1$\times$128 conv}\\ 
\scriptsize{1$\times$1$\times$64 conv}\\ 
\scriptsize{1$\times$1$\times$1 conv}\end{tabular}}                                                                                                                                                                                                                                                                                              & \begin{tabular}[c]{@{}l@{}}
\scriptsize{N$\times$N$\times$128 conv}\\ 
\scriptsize{1$\times$1$\times$64 conv}\\ 
\scriptsize{1$\times$1$\times$1 conv}\end{tabular}                        \\ \hline
\end{tabular}
\end{center}
\end{table}

\subsection{Auxiliary Tasks Prediction}
Due to the complex real-world environments such as the extremely irregular geometric shape of text as well as the complicated image background, the binary information of the license plate is often noisy. Although we intend $G_D$ and $G_R$ to capture robust features for image recovery, the results by this structure do not always guarantee a well-enhanced output. Therefore, our work involves an additional learning branch where a richer feature representation is obtained from the backbone network. Motivated by multi-task learning \cite{caruana1997multitask}, we employ the auxiliary tasks, \textit{i.e.}, binary segmentation and count estimation, which will contribute our main task networks produce more discriminative feature representations. Towards this problem, we sum the weights of the last layer of encoders in order to guide auxiliary task networks to help main task networks effectively extract critical information from the low-quality image.

For the binary segmentation task, we introduce the segment decoder $D_s$ based on U-Net architecture. Detailed architectures of the $D_s$ are shown in Table 1. The $D_s$ accepts feature set $F$ summed from the last features of each main task's encoder and outputs a license plate segment with values indicating the probability of pixels belonging to the license plate. Also, ground-truth labels for segmentation can be inferred from the dotted annotations by \cite{cai2014new}'s method as Otsu Thresholding, as shown in Figure 3. Although our segmentation annotations by \cite{cai2014new} do not fully reflect the actual detail appearance of an image, we have shown in the experiments that this auxiliary  and straightforward learning strategy leads to effective advances in image recovery. Given a pair of $F$ and the ground-truth segmentation result in $I^{seg}$, loss function for the $D_s$ is the binary cross-entropy loss:
    	\begin{equation}
    	    \begin{split}
        	    \mathcal{L}_{D_s}(w) =& \frac{1}{N}\sum_{(x,y) \in N}^{N} I^{seg}_{(x,y)}log(D_{s_w}(F)_{(x,y)})\\ 
        	    & + (1 - I^{seg}_{(x,y)})log(1 - D_{s_w}(F)_{(x,y)}),
    	    \end{split}
    	\end{equation}
    where $I^{seg}_{(x,y)}$ $\in$ $\{0,1\}$ is the real classes of pixels in $I^{seg}$ with 1 for the license plate area and 0 for the background, $D_s(F)_{(x,y)}$ denotes the pixel-wise probability by $D_s$.

Also, we find that the generated recovery samples cannot usually distinguish successive texts due to close to each other. Motivated by the observations, we add a counting decoder $D_c$, which predicts the number of characters in the image. As a result, our $D_c$ plays two roles, where the first is to cause separation between adjacent texts more clearly. The other role is to promote the encoders of each main task to generate a higher quality image while backpropagating the penalty. The loss function for the $D_c$ is the L2 loss:
    	\begin{equation}
    	\begin{split}
        	\mathcal{L}_{D_c} = {\parallel C_{pred} - C_{G.T} \parallel^2},
    	\end{split}
    	\end{equation}
    	where $C_{pred}$ and $C_{G.T}$ are the predicted value and the ground-truth, respectively.

\subsection{Network Training}
The full objective function is a weighted sum of all the losses from Eq. (1) to (4):
    	\begin{equation}
    	\begin{split}
        	\mathcal{L} = { \lambda_{G_D}\mathcal{L}_{G_D} + \lambda_{G_R}\mathcal{L}_{G_R} + \lambda_{D_s}\mathcal{L}_{D_s} + \lambda_{D_c}\mathcal{L}_{D_C}}
    	\end{split}
    	\end{equation} We employ a stage-wise training strategy to optimize main tasks with auxiliary tasks and empirically set the weights of each loss as detailed in Section 5.3.

\subsection{Testing}
At the testing phase, the auxiliary tasks are removed. Given a low-quality test image $I_{test}$, $G_D$ and $G_R$ output the recovered image via denoising and rectification. Then LPR network $LPR$ based on a YOLO v3 detector \cite{redmon2018yolov3} by pre-trained on ImageNet \cite{deng2009imagenet} takes the recovered image and generates the recognition result $LPR_{result}$ of $I_{test}$, and it is denoted as Eq. (6):

    	\begin{equation}
    	\begin{split}
        	{LPR_{result}} = {LPR(G_R(G_D(I_{test})))}.
    	\end{split}
    	\end{equation}

    \begin{figure*}[t]
    \begin{center}
       \includegraphics[width=1.0\linewidth]{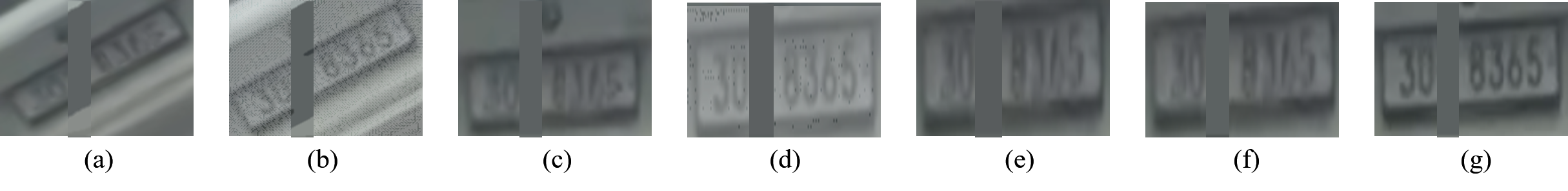}
    \end{center}
       \caption{Ablation Study. (a) : shows noisy input; (b) : only contains denoising net; (c) : only contains rectification net; (d) : adds all main tasks; (e) : adds segment task from (d); (f) : adds counting task from (d); (g) : adds all of tasks, namely our proposed model. }
    \label{fig:long}
    \label{fig:onecol}
    \end{figure*}

\section{Experimental Setting}
In this section, we describe a list of datasets, metric, and implementation details for the proposed method.

\subsection{Datasets}
We use LPR datasets AOLP \cite{hsu2012application} and newly collected dataset, named VTLP. 

\textbf{AOLP-RP} : AOLP-RP \cite{hsu2012application} consists of 611 images collected in Taiwan, including ten numbers and 25 letters (except "O"). This dataset has a challenging factor that the angle of the LP contains oblique samples in terms of distortion. On the other hand, in terms of resolution, all images are relatively easy because they consist of high-resolution samples rather than other datasets.

\textbf{VTLP} : We introduce a new challenging large-scale dataset collected in South Korea. The dataset contains 10,650 LP images, which are divided into 6,400/4,250 images for training and testing, respectively. All Korean letters are hidden for privacy protection. Images in VTLP consist of text(only 10 digits, not Korean). Compared with the public LPR datasets, our dataset has challenging factors: 1) We apply the manual annotation of large-scale images selected from unconstrained real-world, covering a variety of challenging situations using bounding box coordination; 2) Distance from vehicles to the camera is far from other dataset; 3) Various scene-texts interfere with the detection, low-resolution appearance, and very oblique LP.

\subsection{Evaluation Metric}
We follow the evaluation metric that has been widely used in LPR research \cite{hsu2012application,zhuang2018towards}. Therefore, if only one of the consecutive characters is misclassified or not detected, it is treated as a failure case. We denote this metric as a recognition accuracy. Also, we address the 36 characters, including 26 letters and 10 digits for text recognition.

\subsection{Implementation Details}
All the reported implementations are based on the TensorFlow framework, and our method has done on one NVIDIA TITAN X GPU and one Intel Core i7-6700K CPU. In all the experiments, we resize all images to 320 $\times$ 320. For stable training, we use a gradient clipping trick and the Adam optimizer \cite{kingma2014adam} with high momentum. The proposed network is trained in 1 million iterations with a batch size of 16. The weights in all SNIDER layers are initialized from a zero-mean Gaussian distribution with a standard deviation of 0.01, and the constant 0 as the biases in all layers. All models are trained for the first 100 epochs with a learning rate of $10^{-4}$ despite higher values, and then for the remaining epochs at the learning rate of $10^{-5}$. Batch normalization \cite{ioffe2015batch} and LeakyReLU \cite{maas2013rectifier} are used in all layers of our networks. Also, for $LPR$ network as baseline, we use the YOLO v3 detector \cite{redmon2018yolov3} model pre-trained on ImageNet\cite{deng2009imagenet}. 

Two SNIDER models are trained for evaluations and benchmarking with state-of-the-art methods. The first is a backbone model \textbf{SNIDER}, which uses five convolution blocks at encoder and decoder, respectively. In contrast, the other model denoted by \textbf{SNIDER-Tiny} uses a relatively light network thereby too fast for testing. All SNIDER models are trained under the same parameter setting.

    \begin{figure*}[t]
    \begin{center}
       \includegraphics[width=0.8\linewidth]{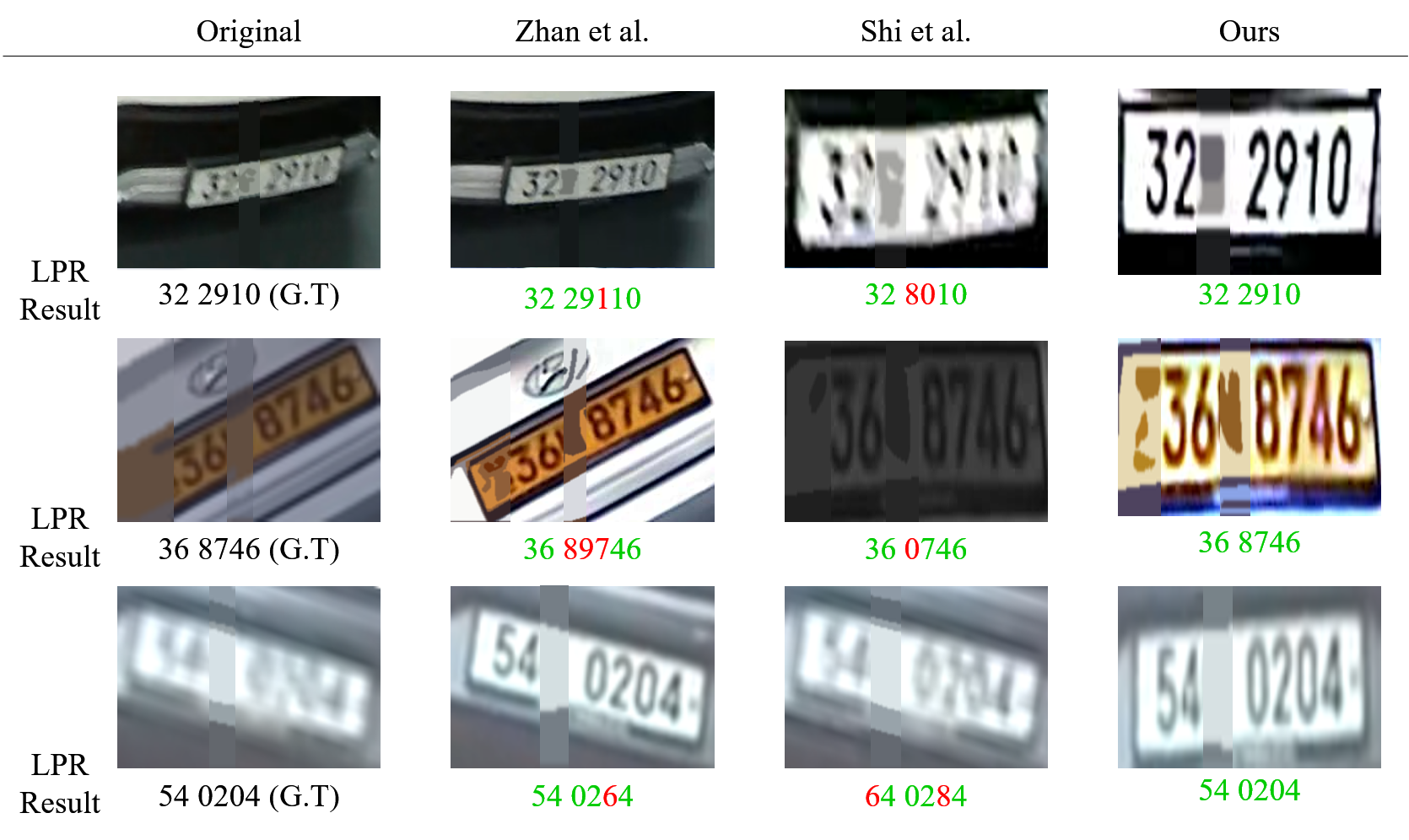}
    \end{center}
       \caption{Visual comparison of different license plate recovery methods: For the three sample images in the first column, columns 2-4 show the recovery images by using \cite{shi2016robust}, \cite{zhang2017beyond} and SNIDER, respectively. The sample images are from VTLP which suffer from geometric distortions as well as low-quality. The proposed SNIDER performs better in LPR recovery. Best viewed on the computer, in color and zoomed in.}
    \label{fig:long}
    \label{fig:onecol}
    \end{figure*}

\section{Results}
In this section, we evaluate the proposed approach on two datasets: AOLP-RP \cite{hsu2012application} and VTLP.

    \begin{table}[]
    \begin{center}
       \caption{Ablation study on the effectiveness of different components. DSN, RSN, SD, and CD represent the $G_D$, $G_R$, $D_s$, and $D_c$, respectively.}
        \begin{tabular}{c|l|cc}
        \hline
        \multirow{2}{*}{Type} & \multirow{2}{*}{Method}     & \multicolumn{2}{c}{LPR Accuracy}   \\ \cline{3-4} 
                              &                             & \multicolumn{1}{c|}{AOLP}  & VTLP  \\ \hline
        a                     & Baseline (YOLO v3)          & \multicolumn{1}{c|}{91.65} & 80.45 \\ \hline
        \multirow{2}{*}{b}    & Add DSN                     & \multicolumn{1}{c|}{91.98} & 84.64 \\
                              & Add RSN                     & \multicolumn{1}{c|}{97.05} & 87.13 \\ \hline
        c                     & Add DSN, RSN                & \multicolumn{1}{c|}{98.53} & 90.71 \\ \hline
        \multirow{2}{*}{d}    & Add DSN, RSN, SD            & \multicolumn{1}{c|}{99.02} & 92.08 \\
                              & Add DSN, RSN, CD            & \multicolumn{1}{c|}{98.69} & 91.08 \\ \hline
        e                     & Add DSN, RSN, SD, CD (ours) & \multicolumn{1}{c|}{\textbf{99.18}} & \textbf{93.08} \\ \hline
        \end{tabular}
    \end{center} 
    \end{table}

\subsection{Ablation Study}
We first compare our proposed method with the baseline LPR network to prove the effectiveness of image recovery performance. Both LPR results on two datasets are reported for the following five types of our methods where each module is optionally added: a) the only baseline without proposed method; b) adding one main task; c) adding all main tasks; d) adding all main tasks and one auxiliary task; e) adding all of the modules (namely, proposed method).

We present the LPR accuracy for each type on two datasets in Table 2, and the visual comparisons are shown in Figure 4. From Table 2, we can find that adding the denoising and the rectification task, respectively, significantly improves the LPR performance (type b, c). In addition, we observe that LPR performance improves more when both tasks are applied at the same time. As shown in Figure 4. (c), noise and blurring effect are removed from the low-quality image (a), and characters are enhanced well compared to (c). This confirms that performing two tasks at the same time is more helpful to recover high-quality images. Despite showing better LPR performance (Table 2. (c)), we still find that the output image contains elements that interfere with LPR performance. For example, there are still challenges to detect the suitable text region, including a region that is unnecessary for recognition, such as a manufacturer's logo (see in Figure 4. (d)), and ambiguity that not well detected between consecutive characters. Therefore, when each auxiliary task is added to main tasks, recovered image quality can be better (Figure 4. (e,f)) and we observe some improvements on LPR performance (Table 2. (d)). Finally, we incorporate all the tasks, perform experiments on it and observe the best performance improvement in LPR (Table 2. (e)). Furthermore, the recovered image in Figure 4. (g) is the most realistic of all results.

   \begin{table}[]
   \begin{center}
   \caption{Full LPR performance (percentage) comparison of our method with the existing methods on \textbf{AOLP-RP \cite{hsu2012application}.}}
        \begin{tabular}{l|c}
        \hline
        \multicolumn{1}{c|}{Method} & AOLP-RP Full LPR accuracy (\%) \\ \hline \hline
        Baseline (YOLO v3)  & 91.65                 \\
        Hsu \textit{et al.} \cite{hsu2012application}  & 85.76  \\
        Li \textit{et al.} \cite{li2018toward}         & 88.38  \\
        Silva \textit{et al.} \cite{silva2018license}  & 98.36  \\
        Zhuang \textit{et al.} \cite{zhuang2018towards}  & 99.02  \\\hline
        SNIDER-Tiny                                & \textbf{98.85}                \\
        SNIDER                                & \textbf{99.18}                \\ \hline
        \end{tabular}
    \end{center}
    \end{table}
    
   \begin{table}[]
   \vspace{-0.0cm}
   \begin{center}
   \caption{Full LPR performance (percentage) comparison of our method with the existing methods on \textbf{VTLP}.}
        \begin{tabular}{l|c}
        \hline
        \multicolumn{1}{c|}{Method} & VTLP Full LPR accuracy (\%) \\ \hline \hline
        Baseline (YOLO v3)  & 80.45                 \\
        Laroca \textit{et al.} \cite{laroca2018robust}  & 87.34  \\
        Silva \textit{et al.} \cite{silva2018license}         & 84.73  \\ \hline
        SNIDER-Tiny  & 86.66  \\ 
        SNIDER                                & \textbf{93.08}                \\ \hline
        \end{tabular}
    \end{center}
    \end{table}

\subsection{Comparison with State-of-the-art Methods}
We compare the proposed method with some state-of-the-art LPR methods \cite{hsu2012application,li2018toward,silva2018license,zhuang2018towards}. For the baseline LPR, the SNIDER has been evaluated over the two datasets as described in Section 4.1 that contain low-quality license plate images with a variety of geometric variations.

As Table 2, 3 and 4 show, the \textbf{SNIDER} consistently outperforms the \textbf{SNIDER-Tiny} across all datasets due to the use of a more in-depth and broader backbone network. However, \textbf{SNIDER-Tiny} is also evaluated to be more effective than most methods, and if not, it shows a relatively small performance difference. Therefore, it can be explained that SNIDER is more useful for LPR than other methods for the low-quality image.

\textbf{AOLP-RP dataset results.} For the AOLP-RP, SNIDER demonstrates that our recovery image can significantly improve the performance of LPR on real-world images. This is mainly due to the fact that AOLP dataset which usually have geometrically tilted cases is processed into a well-rectified image. The results are listed in Table 3, and our method obtains the highest performance \textbf{(99.18\%)}, and outperforms the state-of-the-art LPR methods by more than 0.16\%. Note that what we want to illustrate in the AOLP-RP evaluation (especially see the difference between Baseline and ours in Table 3) is that our method can benefit from the SNIDER, which enhances the image quality despite oblique angle.

\textbf{VTLP dataset results.} The quantitative results for VTLP dataset are shown in Table 4 and the visual comparisons are illustrated in Figure 5. Our approach shows superior performance to other LPR algorithms on LPR accuracy and image recovery. Furthermore, we achieve comparable results with state-of-the-art LPR method \cite{laroca2018robust,silva2018license}. From Table 4, our method obtains the highest performance \textbf{(93.08\%)}, and outperforms the state-of-the-art methods by more than 5.74\% (87.34\% \textit{vs} 93.08\%). Note that SNIDER achieves robust performance in VTLP that are collected in low-resolution environments rather than other datasets.

    \begin{figure*}[t]
    \begin{center}
       \includegraphics[width=0.8\linewidth]{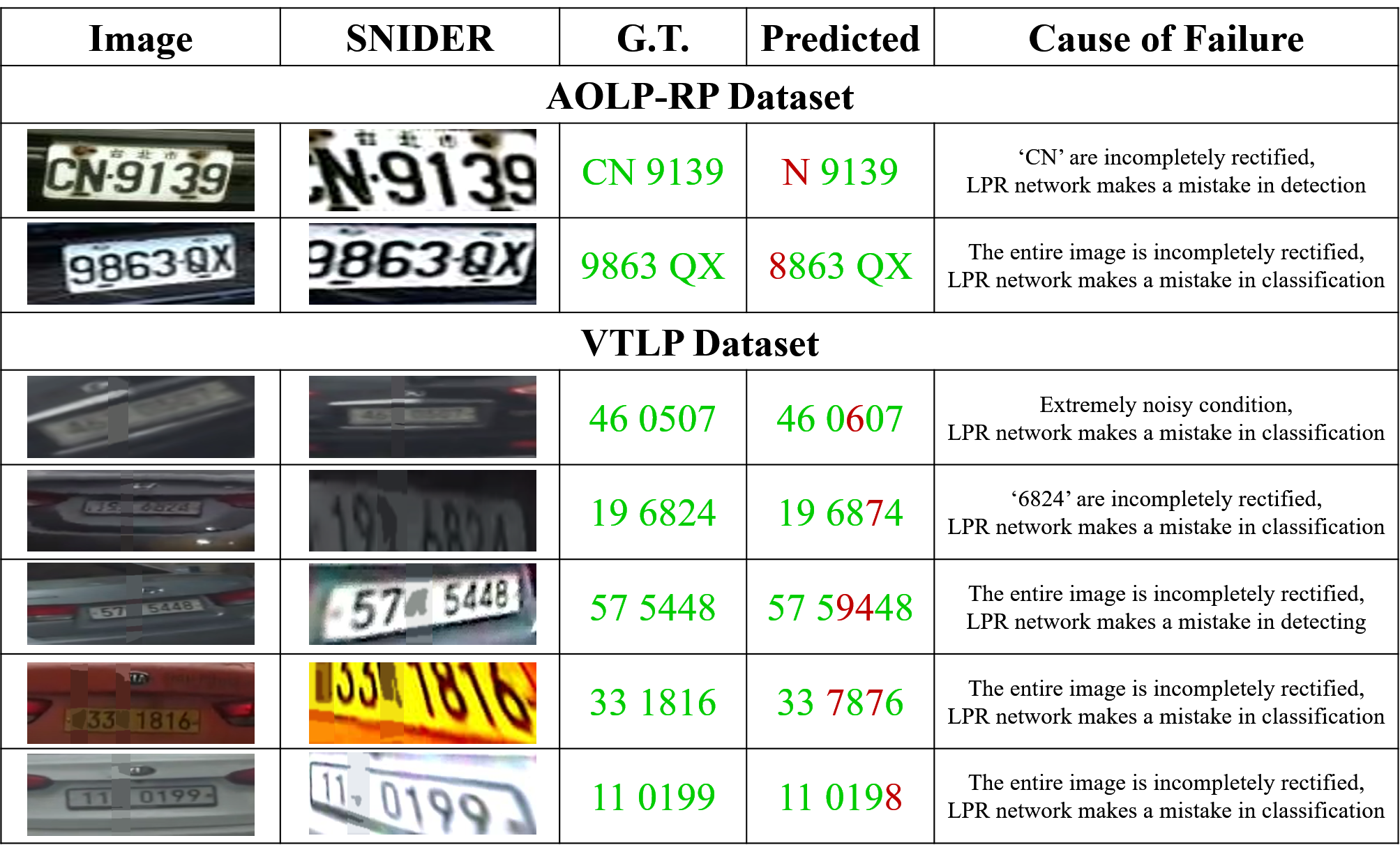}
    \end{center}
       \caption{Error study on AOLP and VTLP dataset. Best viewed on the computer, in color and zoomed in.}
    \label{fig:long}
    \label{fig:onecol}
    \end{figure*}

\subsection{Parameter Study of the Weights for Tasks}
The set of weights $\lambda$ in Eq.(5) determines the influence of each task. To choose the optimal selection of $\lambda$, we perform various experiments with the SNIDER model on AOLP-RP and VTLP dataset. Since the influence of the main task is larger than that of the auxiliary task, the weight is also set higher. We also need to adjust the weights for fast optimization even within the auxiliary task. Figure 4 shows the segment decoder $D_s$ plays an important role in eliminating unnecessary areas that interfere with LPR. Therefore, we set the weight of the segment decoder higher than the counting decoder. In our experiment, we set the weights for $\lambda_{G_{D}}$, $\lambda_{G_{R}}$, $\lambda_{D{s}}$ and $\lambda_{D{c}}$ to 0.4, 0.4, 0.15, and 0.05, respectively.

\begin{table}[]
\begin{center}
   \caption{Impact of improving the LPR network and its performance evaluation with SNIDER for the VTLP testing set.}
\begin{tabular}{l|c|c}
\hline
Baseline            & LPR Accuracy & FPS  \\ \hline
Faster R-CNN \cite{ren2015faster}        & 87.06             & 2.7  \\
CornerNet-Squeeze \cite{law2019cornernet}  & 93.39             & 13.1 \\
CenterNet ResNet-18 \cite{zhou2019objects} & 84.68            & 46   \\ \hline
YOLO v3 \cite{redmon2018yolov3} (SNIDER-Tiny)             & 86.66             & 44   \\
YOLO v3 \cite{redmon2018yolov3} (ours)             & 93.08             & 37   \\ \hline
\end{tabular}
\end{center}
\end{table}

\subsection{Impact of LPR Network}
We evaluate how LPR network choice impact LPR performance on the VTLP testing set. Results are shown in Table 5. We mainly adopt a real-time detector for fast processing. Compare with \cite{ren2015faster,zhou2019objects}, SNIDER indicates that the detector plays an important role in LPR performance. Although previous detectors are high-speed processing through lightweight models, they do not guarantee accuracy. Thus, we adopt YOLO v3, which corresponds to the adequate model that includes enough capacity for rich feature representation during real-time processing.

\subsection{Weakness Analysis}
 Figure 6 shows some failure cases, including some false recovery results. These results identify that more progress is needed to improve the rectification performance further. Future work will address this problem by adding the adjacent context to recovering these more challenging license plate images.

\section{Conclusion}
 In this paper, we propose a new end-to-end trainable image recovery method that is capable of recognizing license plates in the real-world. The proposed recovery network consists of two sub-networks, the denoising sub-network and the rectification network. In particular, two auxiliary tasks are designed to leverage the recovery of license plates, promoting the feature set to be more robust against the geometric variations and blurry data in the real-world scenes. Moreover, a new loss function is introduced to the backbone network to provide regularization effects and a higher-quality recovery image. Extensive experiments over various datasets demonstrate superior performance in license plate recovery and recognition. 

\section*{Acknowledgement}
This work was supported by Institute of Information \& communications Technology Planning \& Evaluation (IITP) grant funded by the Korean government (MSIT) (No.B0101-16-0525, development of global multi-target tracking and event prediction techniques based on real-time large-scale video analysis. We also appreciate useful discussions with Kyungho Won, Jaewoong Yun, and Sangwoo Park.

{
\bibliographystyle{ieee}
\bibliography{egbib}
}

\end{document}